\documentclass{article}

\usepackage[nonatbib, final]{neurips_2023}

\usepackage[utf8]{inputenc} 
\usepackage[T1]{fontenc}    
\usepackage{hyperref}       
\usepackage{url}            
\usepackage{booktabs}       
\usepackage{amsfonts}       
\usepackage{nicefrac}       
\usepackage{microtype}      
\usepackage{xcolor}         
\usepackage{graphicx}
\usepackage{csquotes}
\usepackage[exponent-product=\times, retain-unity-mantissa=false]{siunitx}  
\usepackage{amsmath}
\usepackage{caption}
\usepackage{subcaption}
\usepackage{multirow}

\title{A Generative Model for Accelerated Inverse Modeling Using a Novel Embedding for Continuous Variables}

\author{
    S\'ebastien Bompas $^1$ \;and\; Stefan Sandfeld$^{1,2}$\\
    $^1$Institute for Advanced Simulations: Materials Data Science and Informatics (IAS-9)\\
    Forschungszentrum J\"ulich GmbH \\
    52425 J\"ulich, Germany\\
    $^2$Faculty of Georesources and Materials Engineering,\\
    RWTH Aachen University\\ 
    52056 Aachen, Germany\\
    \texttt{\{s.bompas, s.sandfeld\}@fz-juelich.de} \\
}
	
\begin{document}

\maketitle

\begin{abstract}
    In materials science, the challenge of rapid prototyping materials with desired properties often involves extensive experimentation to find suitable microstructures. Additionally, finding microstructures for given properties is typically an ill-posed problem where multiple solutions may exist.     
    Using generative machine learning models can be a viable solution which also reduces the computational cost. This comes with new challenges because, e.g., a continuous property variable as conditioning input to the model is required. 
    We investigate the shortcomings of an existing method and compare this to a novel embedding strategy for generative models that is based on the binary representation of floating point numbers. 
    This eliminates the need for normalization, preserves information, and creates a versatile embedding space for conditioning the generative model. This technique can be applied to condition a network on any number, to provide fine control over generated microstructure images, thereby contributing to accelerated materials design.
\end{abstract}

\section{Introduction}
    Machine learning in the domain of material science is a quickly growing field and can provide significant benefits for designing new material, optimizing material for specific applications, or enabling fast characterization \cite{Chen2021, singh2018physicsaware}. Generative models such as generative adversarial networks (GANs) \cite{GAN} have already been successfully applied for accelerating materials discovery \cite{deepgen, Microstructuresynthesis, MMD}. 
    Enforcing the desired material property in the generated images can be a challenging task and requires careful design of the architecture in order to get the desired output. 
    Most methods use class-based inputs, which work well for categorical data but may not be suitable for materials with desired properties (i.e., continuous values). Approaches, such as the CcGAN \cite{Ccgan}, introduce new strategies to condition the network on continuous values but struggle to generalize to microstructure synthesis requiring precise physical properties.
    Debugging such a complex network is non-trivial and has only rarely been attempted in the literature; to the best of our knowledge, there exists no previous work with direct relevance for the investigated inversion of the structure-property relation. 
    Rich embedding spaces have been shown to play a key role in text-to-image synthesis diversity \cite{voynov2023P+, LWEE, CLIP}. This led us to investigate and analyze the embedding space of a GAN, as it is a good approach to understand the details of the generation process.
    In particular, it is found that for our problem the latent space of a CcGAN lacks diversity: more than a third of the neurons appear to be dead. In addition, details of the embedding space are to a large extent determined by the weight initialization at the beginning of the training. As a remedy for this issue and to enable full control over the generated microstructure, we propose a novel and somewhat unusual embedding strategy that is based on the binary representation of floating point numbers.

\section{Method}

    \textbf{Conditioning Generative Models.}\;    
        When conditioning networks, we typically provide extra information via labels, which is straightforward for discrete, class-based inputs via one-hot encoding or a lookup table with learnable weights \cite{CGAN}. However, in our case, the dataset contains \emph{continuous} physical quantities tied to microstructures.       
        One could simply bin those values and use existing conditioning methods but this has some drawbacks: the number of bins or classes will define the diversity of the generated images as the bins represent a range of values. This embedding strategy will be used as a baseline, and we will also evaluate the impact of the number of classes on the overall performance.  
        To address this, \cite{Ccgan} introduced new loss terms and an improved label input that utilizes an autoencoder to map normalized floats to an embedding space for conditioning.
        Although this approach gives reasonable results in some cases, we found that it fails to generalize to microstructure synthesis where an exact physical property is wanted or required as going from a single number to a higher embedding space is not a trivial task and is strongly dependent on the weights initialization of the autoencoder.
        To keep the precision that we get from the continuous value and the ability to control the output of the GAN, we introduce a new way of conditioning the GAN on those continuous values. 

\begin{figure}[!htb]
            \centering
            \includegraphics[width=0.95\textwidth]{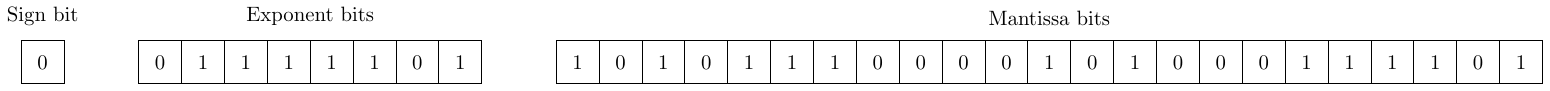}
            \caption[]{Binary representation example of a float32 number. Here, the number $0.42$ is shown in binary representation, with its three components clearly separated.}	
            \label{fig:binary_example}
        \end{figure}

    \textbf{Binary Representation of Floating Point Numbers.}\;
        Floating point numbers are stored in memory as an array of bits, and the precision used to represent that number determines the number of bits required to store the value. 
        The IEEE 754 standard defines the details of this representation \cite{IEEE754}. The standard representation of a single precision number (also referred to as \enquote{float 32}) uses 32 bits for the
        representation of the value: one bit is used to store the \emph{sign}, 8 bits are used to store the \emph{exponent}, and 23 bits are used to store the \emph{significand} or \emph{mantissa}. 
        An example of such representation is given in Fig.~\ref{fig:binary_example}. 
        This representation allows a wide dynamic range of values that can be described.
        The maximum value that can be stored in a single precision number is $(2 - 2^{-23})\times2^{127} \approx 3.4 \times 10^{38}$. Other numerical precision, such as double precision or half precision, offer the ability to use different value ranges and have different memory requirements.

    \textbf{Binary Embedding of Continuous Variables.}\;
        To maximize the effectiveness of this representation in conditioning GANs, we can split the resulting vector from the binary representation into its three basic parts: the sign, the exponent, and the mantissa. Each part is then fed to a different fully connected layer before being concatenated back into one vector and fed to a fully connected layer. Doing so allows each part to be learned separately alongside the relationship between the constituting bits before being merged. The resulting embedded representation can then be used to condition the image generation as done in CcGAN, and the architecture is called \emph{binary embedding conditioning GAN} (BcGAN). Moreover, it should be noted that obtaining the binary representation from a float can be achieved via straightforward mathematical operation, and training these simple linear layers is orders of magnitude faster than that of the GAN architecture, thus making this embedding strategy computationally efficient. 
        Appendix~\ref{appendix:model} contains more details on the architecture design.

\section{Details of the Experiment}
    
    The dataset is obtained from simulations of the \emph{Ising model}, a statistical physics model used to describe ferromagnetic materials and phase transitions. It was introduced by Lenz \cite{Lenz} and later solved analytically in one dimension by Ising \cite{Ising1925}. This model is a mathematical representation of the behavior of magnetic spins in a crystalline lattice, describing the interaction between neighboring spins. Here, each spin is represented by a pixel of an image. Each image is the result of a unique simulation with random initial values. See Appendix~\ref{appendix:Ising} for more details of the model.
    The resulting dataset comprises widely different microstructures that are strongly temperature-dependent, and thus, the temperature is the property under consideration. In this work, the Ising dataset serves as a representative  example of a broader class of structure-property relations.

    Standard approaches to assess the quality of GAN-generated images typically rely on metrics such as the inception score \cite{IS} or the Fréchet inception distance \cite{FID} which measure the similarity between two sets of feature vectors. However, these metrics do not provide insights into the \emph{physics-related} accuracy of the generated images, which is particularly important for designing microstructures such as  those obtained from the Ising model.

\begin{figure}[!htb]
        \centering
            \includegraphics[width=\textwidth]{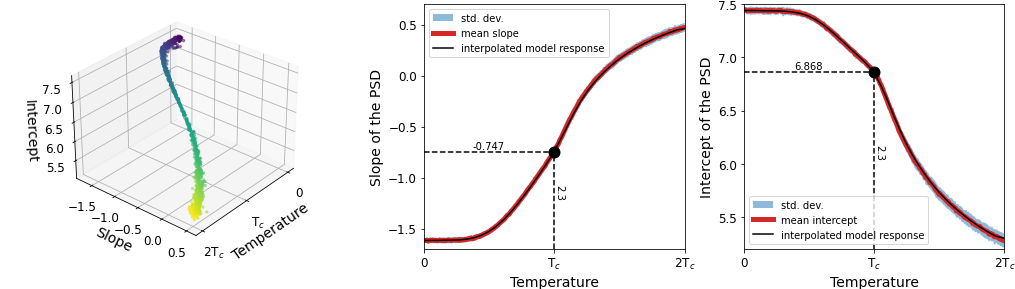}
            \caption{From left to right: 3D representation of the model response and visualization of the interpolated model response. From the slope and intercept of the generated images, we can map them to the corresponding temperature-property.}
            \label{fig:ising:featurespace}
    \end{figure}

    The power spectral density (PSD) is a measure used in various fields to analyze signal power distribution across different frequencies. In image analysis, PSD reflects the distribution of feature sizes within an image and is obtained through Fourier transformations. When applied to the image data, PSD reveals information about the size distribution (e.g., the magnetic domains of the Ising model). It is a means to differentiate microstructures at different temperatures. To simplify the analysis, a linear fit in log-log space is applied to the PSD from which the slope and intercept of the fit serve as a two-parameter representation. 
    These two parameters, obtained from images of the training dataset,  are a function of temperature; they are distributed around a mean with only a small variance (cf. Fig.~\ref{fig:ising:featurespace}) and therefore are suitable as features for identifying microstructures \cite{nguyen2023efficient}. 
    Computing the two values for a generated image allows us to obtain the corresponding property (the temperature). Comparing this to the conditioning temperature is used as a measure for the accuracy of the generated microstructure. By generating multiple images for various temperatures, the GAN's ability to capture underlying microstructures can be evaluated by comparing the temperature of the generated images to the expected temperature used for conditioning. Details regarding the training of each model can be found in the Appendix \ref{appendix:model}.

\section{Results and Discussion}

    \textbf{Model Performances and Parameters.}\;
    After each model was trained, we generated images for temperatures ranging from $0$ and $2\times \mathrm{T}_c$. The two PSD parameters are used to evaluate the accuracy of the generated microstructures.  
    \begin{figure}[!htb]
        \centering
        \begin{subfigure}[b]{0.44\textwidth}
            \includegraphics[width=\textwidth]{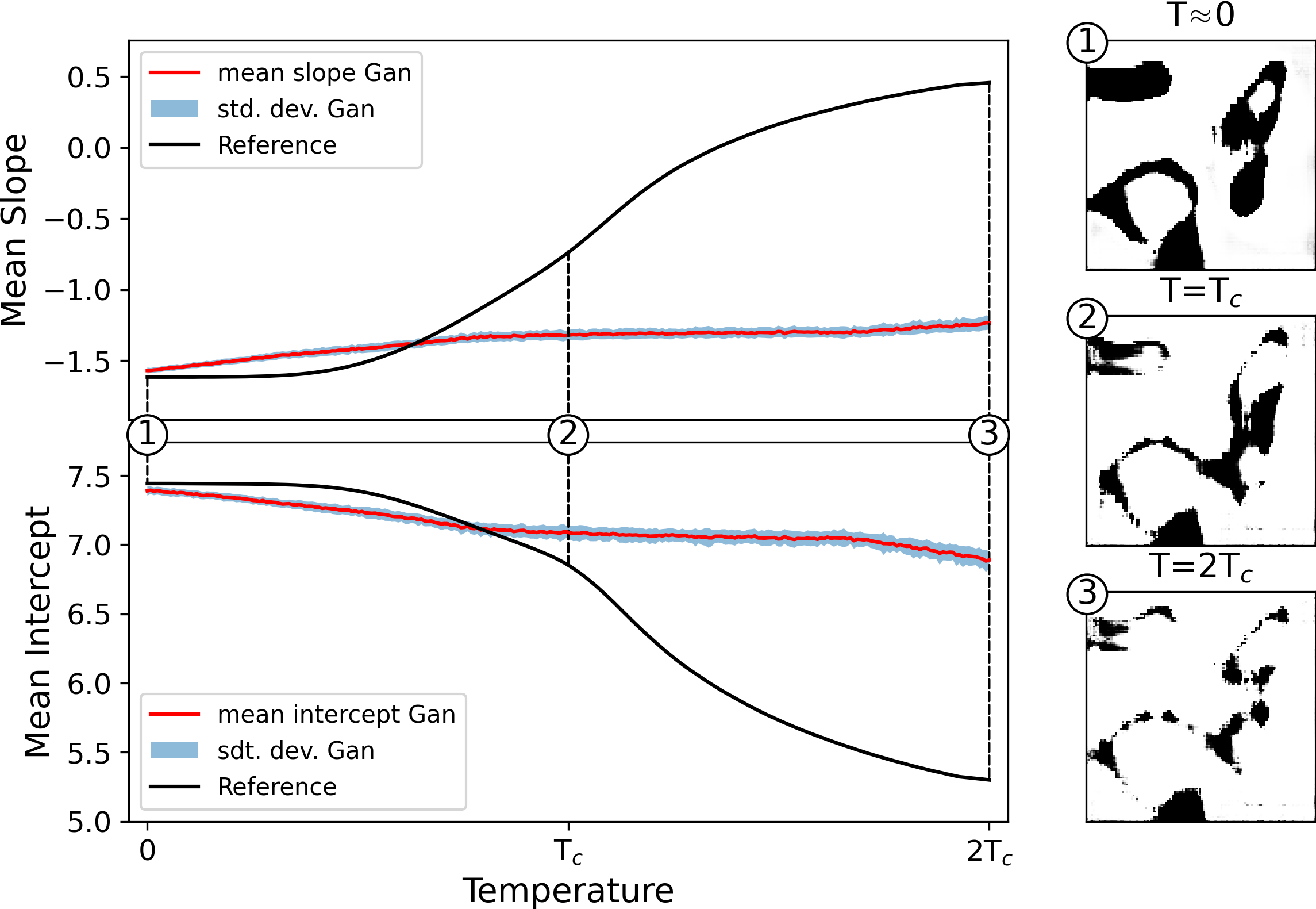}
            \caption[]{CcGAN: the temperatures were uniformly sampled from the normalized temperatures.}	
            \label{fig:psd_Ccgan}
        \end{subfigure}
        \hfill
        \begin{subfigure}[b]{0.44\textwidth}
            \centering
            \includegraphics[width=\textwidth]{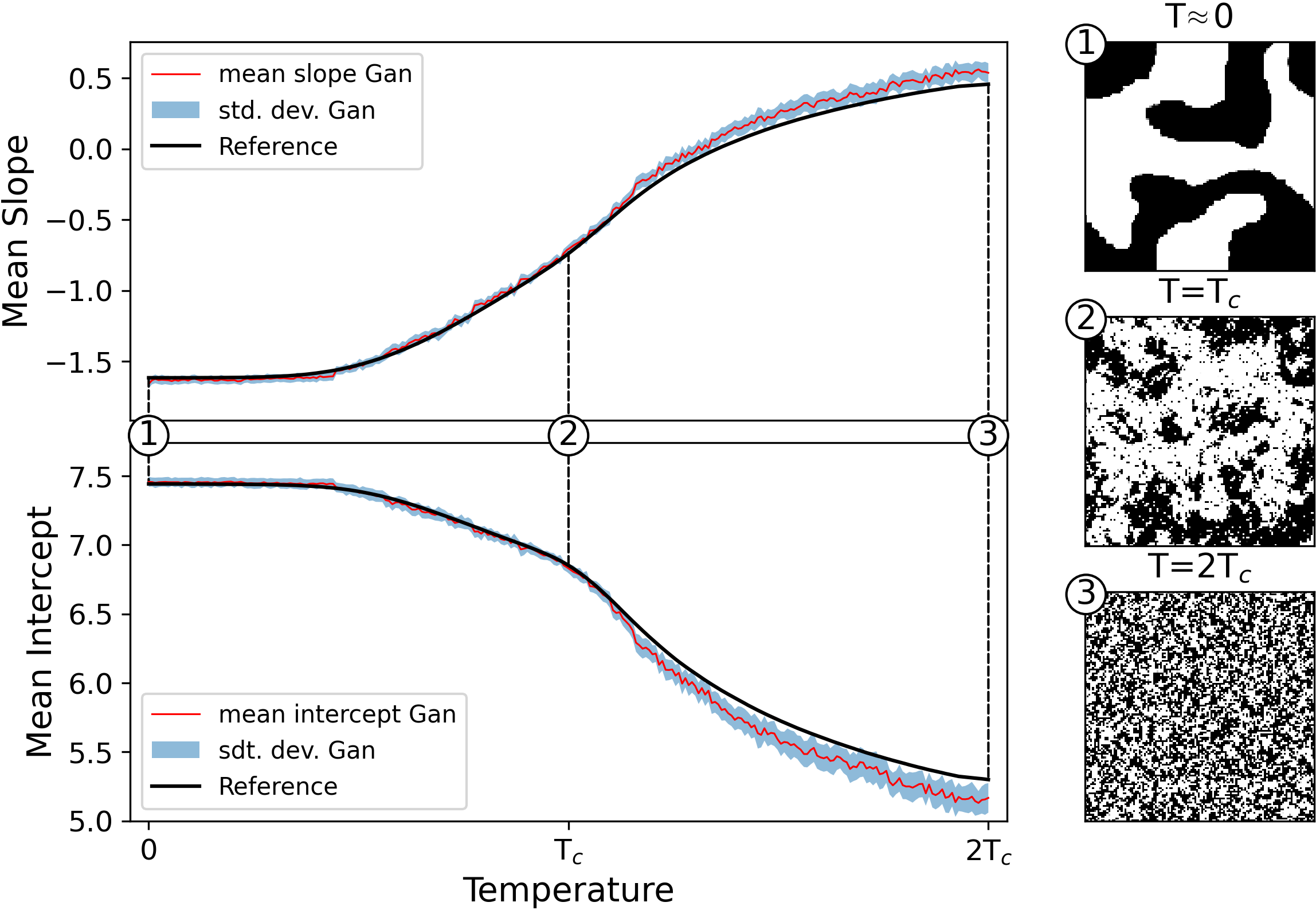}
            \caption[]{binary embedding (BcGAN): the temperature is uniformly sampled between $0$ and $2\times \mathrm{T}_c$.}	
            \label{fig:psd_BCgan}
        \end{subfigure}
        \caption{Comparison of the two PSD parameters and the microstructures for both embedding strategies. Each temperature is sampled 100 times to get an average estimation of the model response.}
        \label{fig:PSDs graphs}
    \end{figure}
    Fig.~\ref{fig:psd_Ccgan} shows that the CcGAN model fails to capture the underlying property of the model material and additionally suffers from mode collapse. When the binary representation is used for the embedding space, the model can reproduce the different characteristics of the microstructure throughout the entire temperature range (see Fig.~\ref{fig:psd_BCgan}). 
    \begin{figure}[!htb]
    	\bigskip
        \centering
        \begin{subfigure}[b]{0.32\textwidth}
            \includegraphics[width=\textwidth]{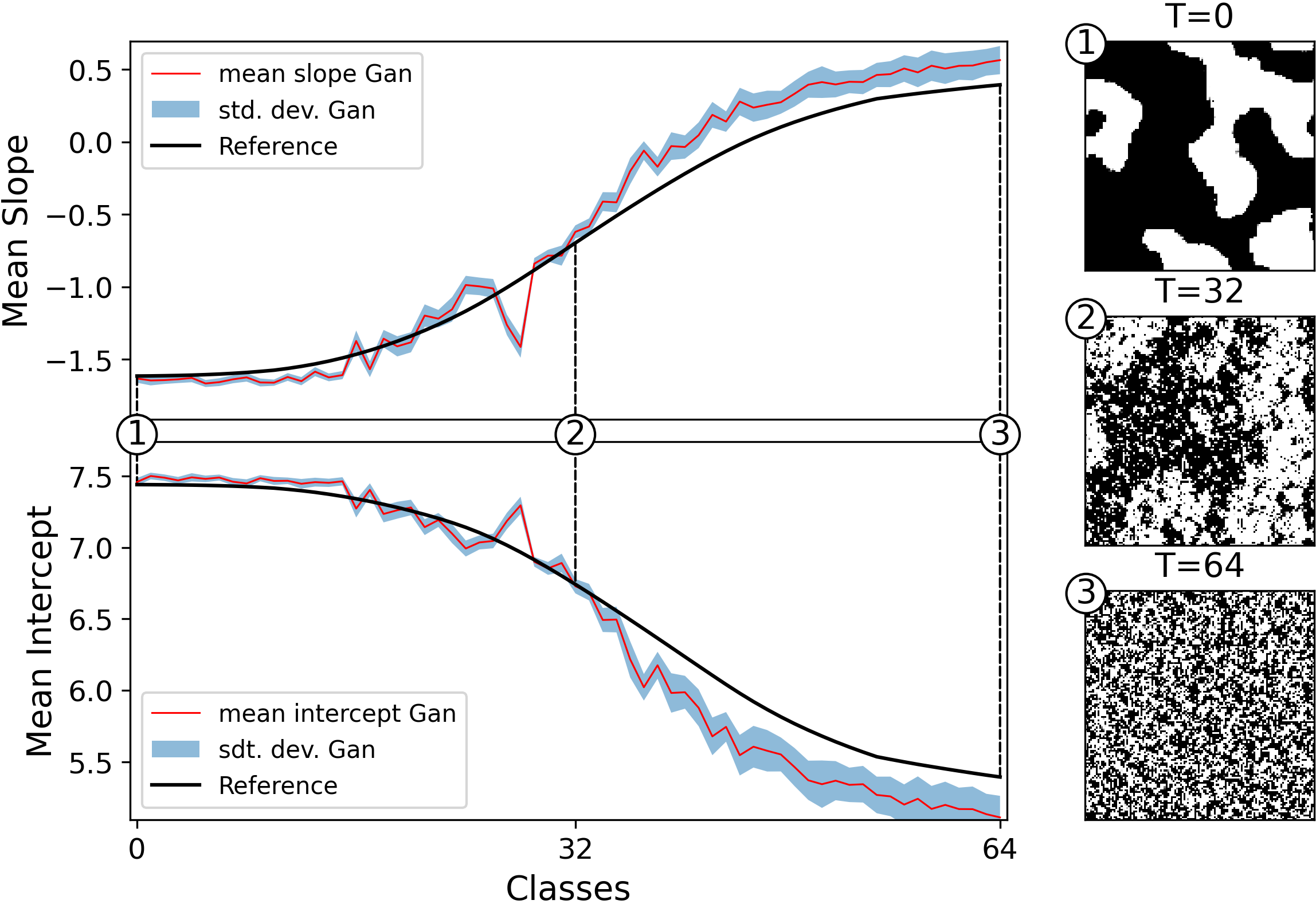}
            \caption[]{CGAN trained with 64 classes.}	
            \label{fig:psd_Cgan_64}
        \end{subfigure}
        \hfill
        \begin{subfigure}[b]{0.32\textwidth}
            \centering
            \includegraphics[width=\textwidth]{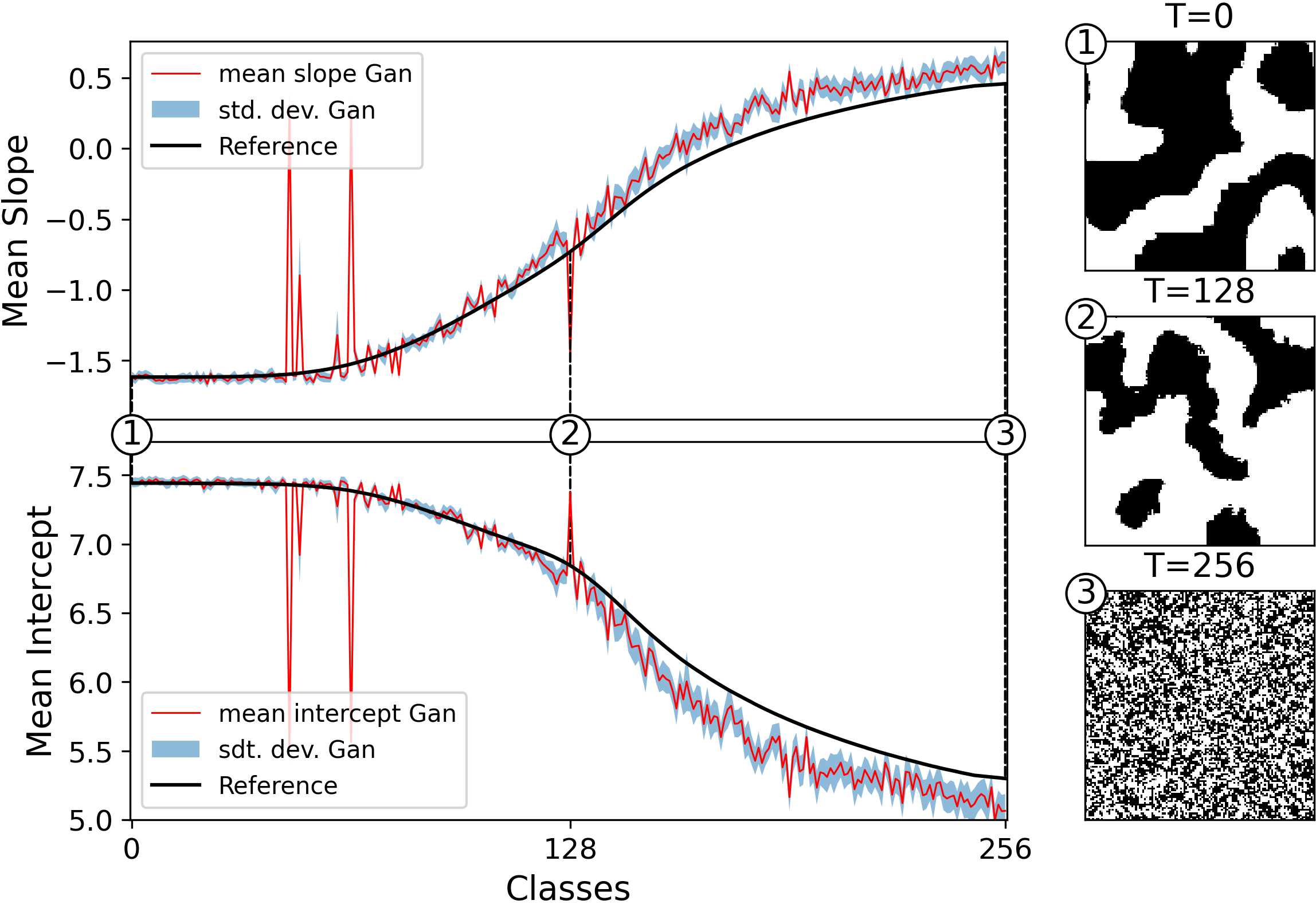}
            \caption[]{CGAN trained with 256 classes.}	
            \label{fig:psd_Cgan_256}
        \end{subfigure}
        \hfill
        \begin{subfigure}[b]{0.32\textwidth}
            \centering
            \includegraphics[width=\textwidth]{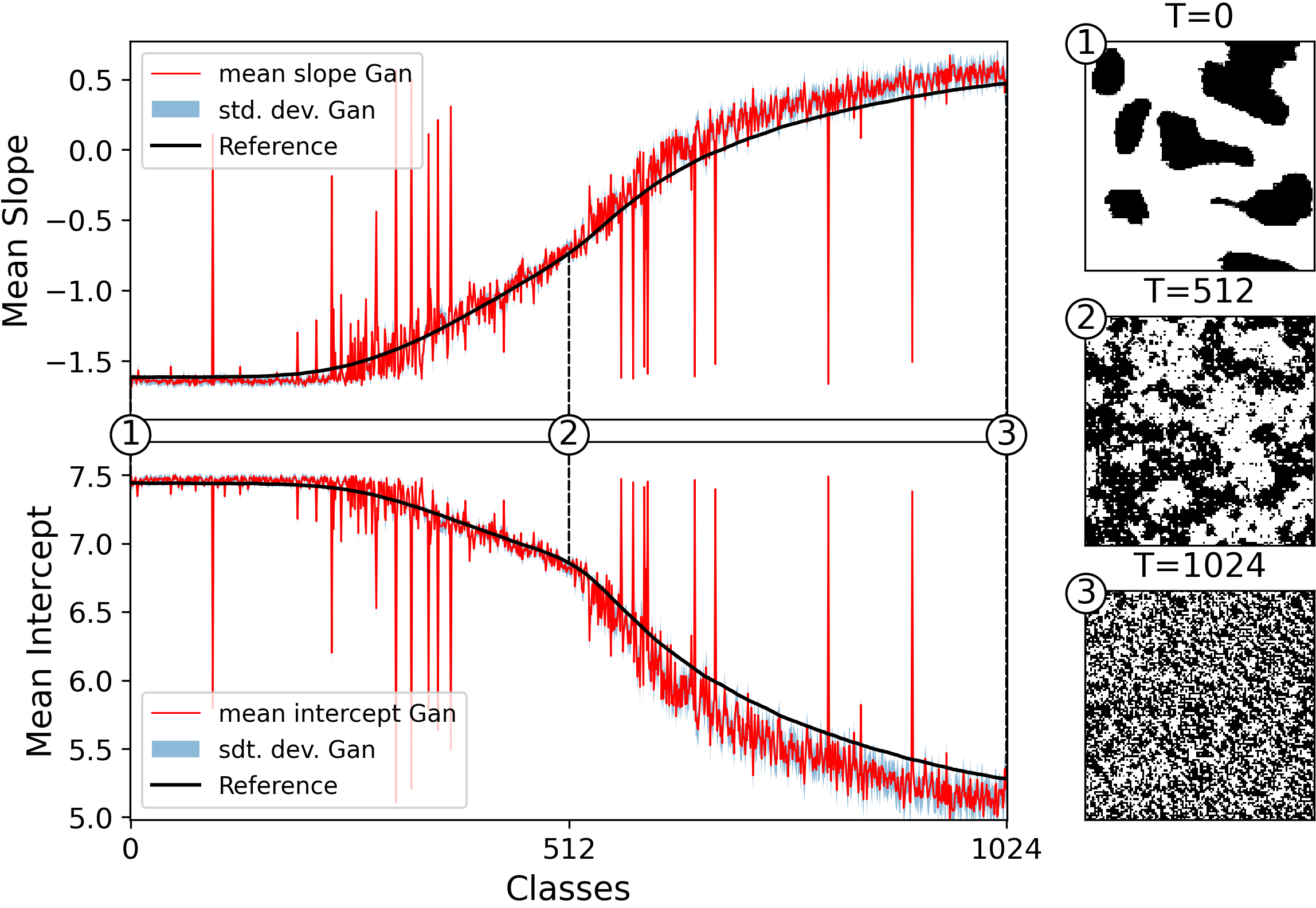}
            \caption[]{CGAN trained with 1024 classes}	
            \label{fig:psd_Cgan_1024}
        \end{subfigure}
        \caption{Comparison of the two PSD parameters and the microstructures for the CGAN with different numbers of classes. Each temperature is sampled 100 times to get an average estimation of the model response. Here, we can clearly see that increasing the number of classes gives increasingly worse results.}
        \label{fig:PSD Cgan}
    \end{figure}

    Although the CGAN embedding strategy gives good results at a small number of classes as shown in Fig.~\ref{fig:psd_Cgan_64}, increasing the number of classes in order to get a finer control of the generated images deteriorates the network performance (Figs.~\ref{fig:psd_Cgan_256} and \ref{fig:psd_Cgan_1024}). In addition, increasing the number of classes has a direct impact on the memory footprint of the network as the number of parameters of embedding drastically increases as shown in Table.~\ref{tab:train&cost}.
    
    \noindent\begin{table}[]
        \centering
        \begin{tabular}{c c c c c c c}
            \toprule
            \textbf{Model type}  & \multicolumn{5}{c}{\textbf{Number of parameters}}                  & \textbf{Training} \\
                                 & Total            & Gen    & Disc   & Embed G       & Embed D       & \textbf{time}\\
            \midrule
            \textbf{CcGAN}       & 75.21M           & 33.48M & 41.60M & 21.61M~(67k)  & 21.61M ~(67k) & 24.1h ~(3.1h) \\
            \midrule
            \textbf{CGAN}        & & & & & & \\
            64 classes           & 14.64M           & 7.23M  & 6.36M  &  8.2K         & 1.05M         & 27.4h\\
            256 classes          & 17.85M           & 7.23M  & 6.36M  & 32.8K         & 4.2M          & 27.5h\\
            1024 classes         & 30.52M           & 7.23M  & 6.36M  &  132K         & 16.8M         & 28.4h\\
            
            \midrule
            \textbf{BcGAN}       & 14.42M           & 6.37M  & 6.57M  & 1.18M         & 308K          & 22.5h\\
            \bottomrule
        \end{tabular}
        \caption{Summary of the computation cost and training time for the different model architectures. For each model, the size of the generator and the discriminator are given without their respective embedding strategy, which is summarized in the following column. For the CcGAN architecture, the conditioning is based on a pre-trained ResNet34 on the dataset and is then used to train a multilayer perceptron (MLP) to reconstruct the labels. The size of the ResNet is listed as part of the embedding but only the MLP (in parenthesis) is counted in the GAN architecture. The training time for the ResNet is listed in parenthesis and is counted for the total training time}
        \label{tab:train&cost}
    \end{table}

    \textbf{Embedding Space Statistics.}\;
    By extracting the embedding layer from each model we can evaluate their latent spaces. We generate a uniformly distributed vector of values in their respective working range (i.e., normalized to [0,1] for the CcGAN, using the original  range of [0, $2\mathrm{T}_c$] for the BcGAN, and [0,255] class bins for the CGAN). This vector is then fed to the corresponding embedding network to monitor the output of each neuron activity based on the input value. The neuron output statistics of the first 10 neurons can be seen in Fig.~\ref{fig:latents graphs}. An untruncated version of this plot can be found in Appendix~\ref{appendix:latent}. 
    \begin{figure}[!htb]
        \centering
        \begin{subfigure}[b]{0.32\textwidth}
            \centering
            \includegraphics[width=\textwidth]{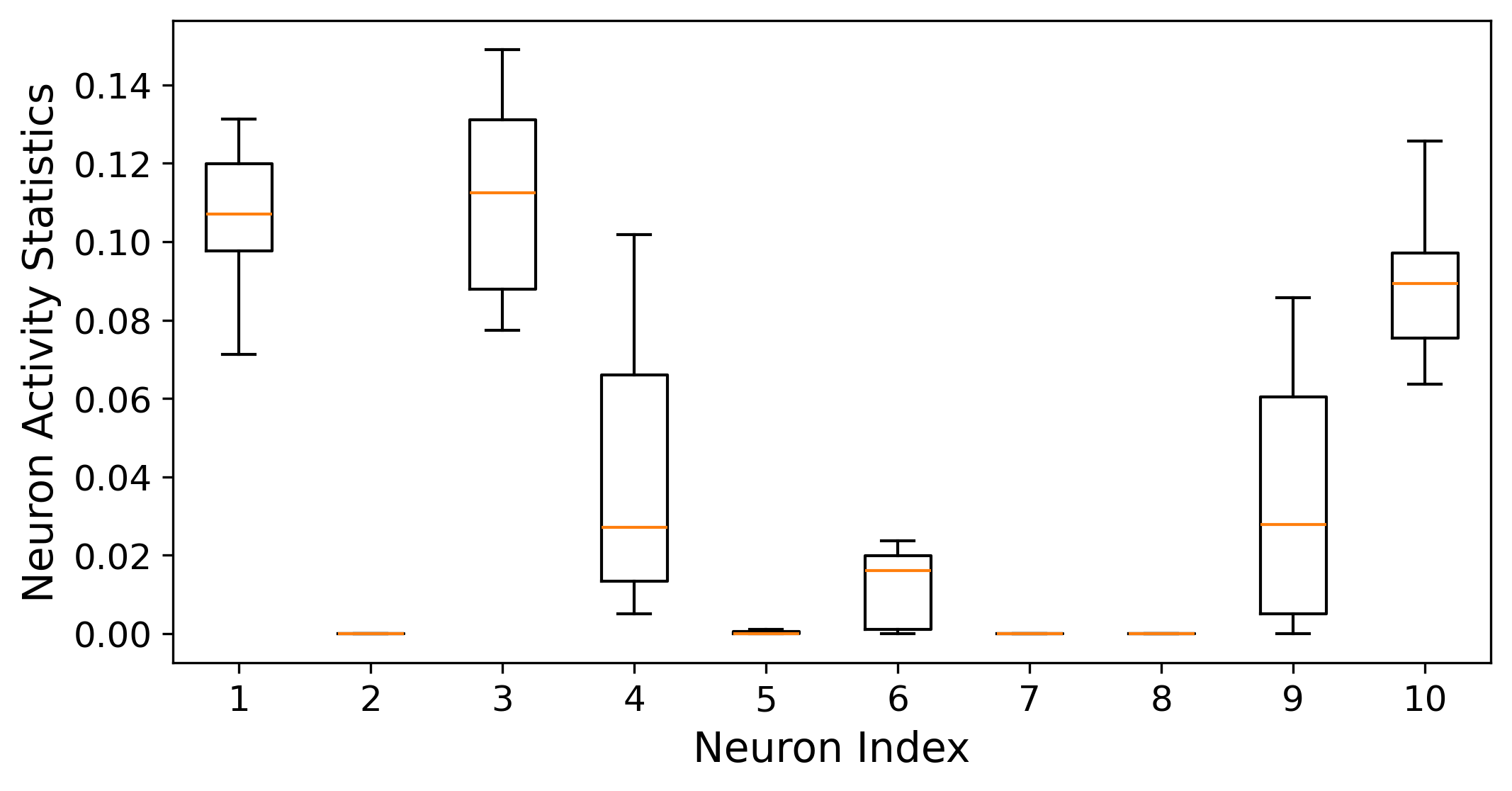}
            \caption{CcGAN embedding space}
            \label{fig:CCGAN_latent}
        \end{subfigure}
        \hfill
        \begin{subfigure}[b]{0.32\textwidth}
            \centering
            \includegraphics[width=\textwidth]{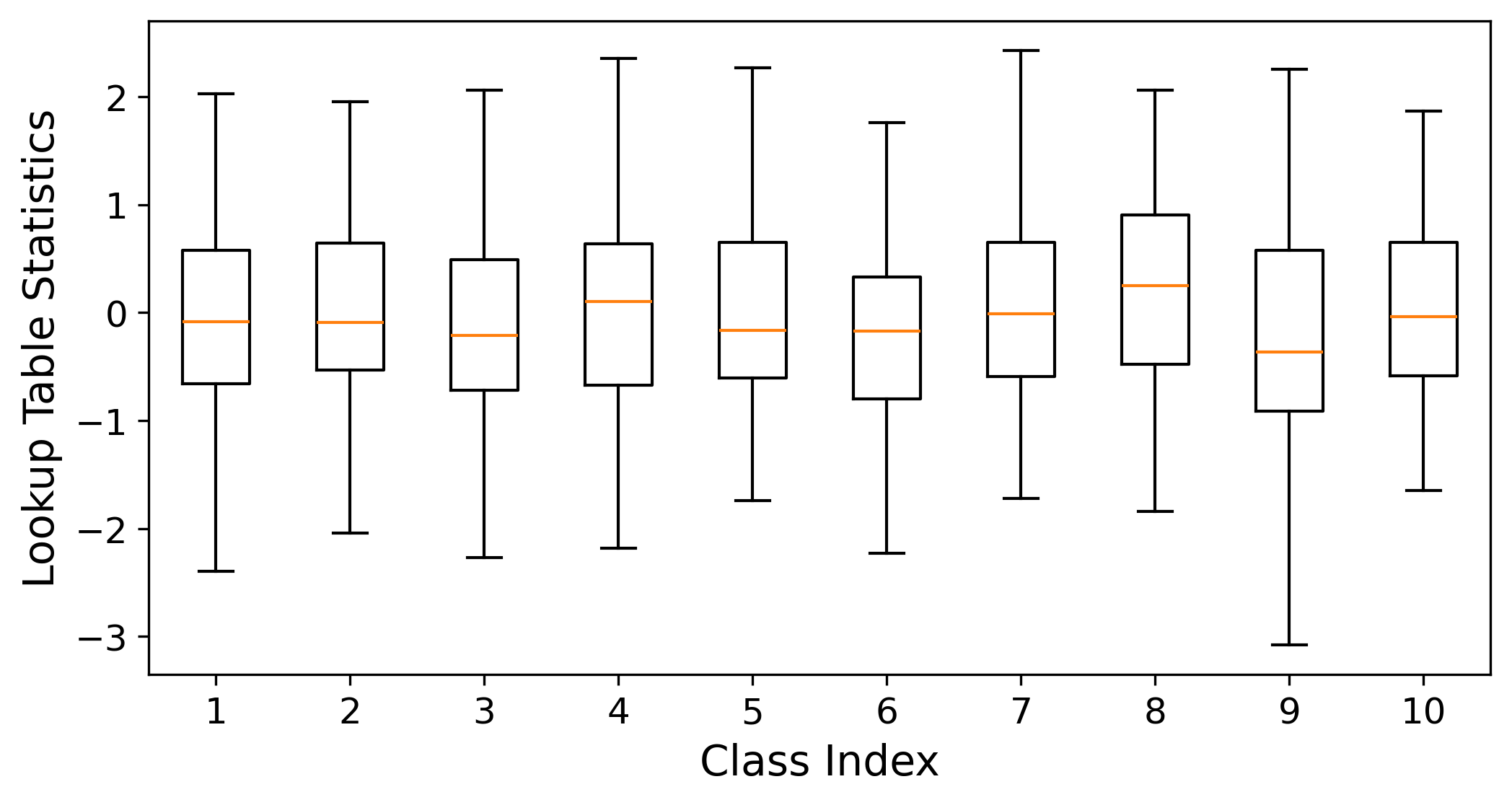}
            \caption{CGAN embedding space}
            \label{fig:CGAN_latent}
        \end{subfigure}
        \hfill
        \begin{subfigure}[b]{0.32\textwidth}
            \centering
            \includegraphics[width=\textwidth]{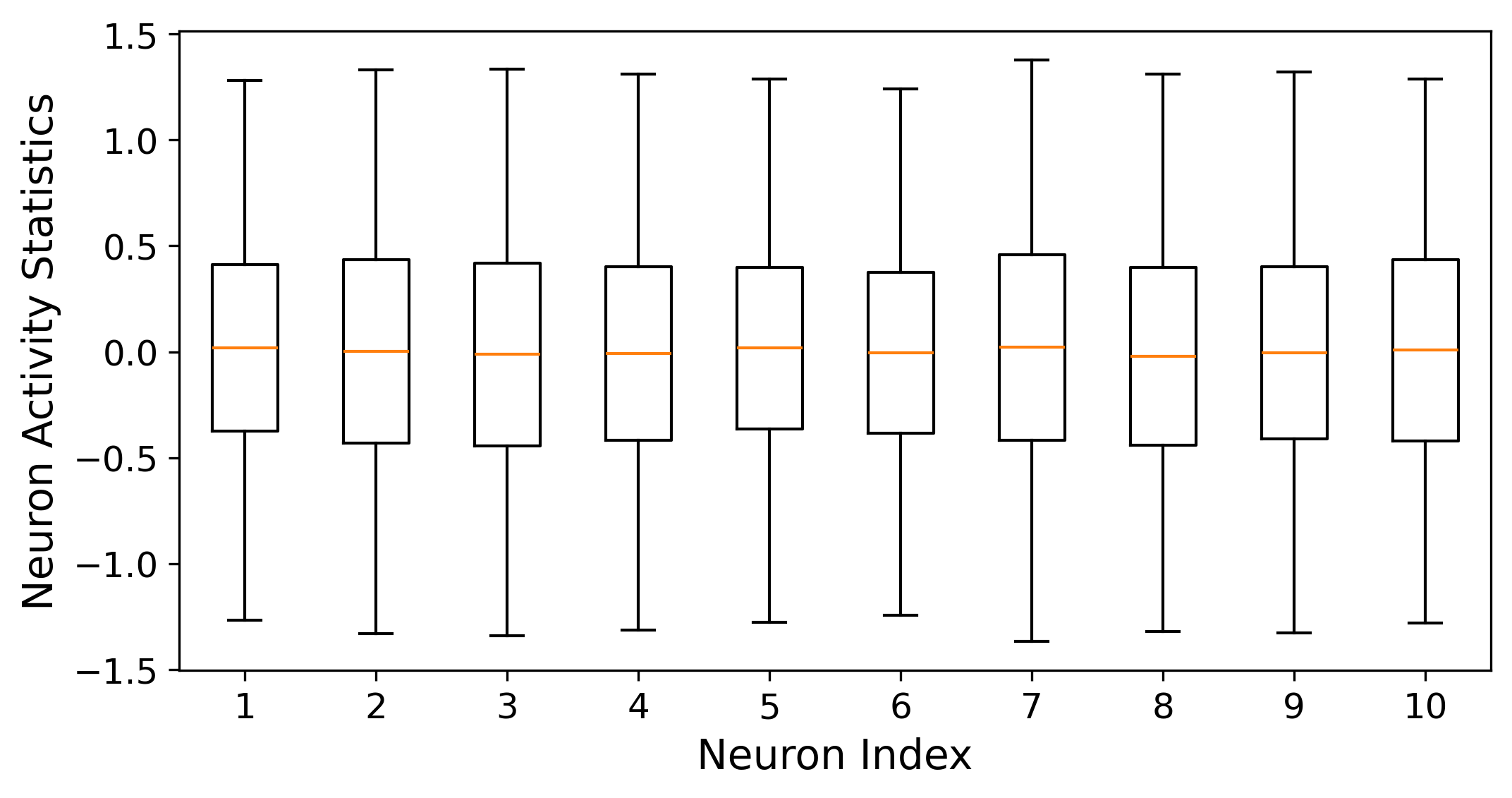}
            \caption{BcGAN embedding space}
            \label{fig:BCGAN_latent}
        \end{subfigure}
        
        \caption{Output activity of the first 10 neurons from the embedding space of each generator model. For the CGAN architecture, the first 10 entries of the lookup table are used for the classes case.}
        \label{fig:latents graphs}
    \end{figure}
    As can  be seen in Fig.~\ref{fig:CCGAN_latent}, a significant number of neurons do not respond to any inputs and must be considered as dead neurons. This has the effect that the latent space has regions that contain no significant information to condition the generated images. A detailed study of the complete CcGAN latent space shows that more than 35\% of the neurons are dead neurons. This percentage stays roughly the same even when the number of neurons is changed. Furthermore, the output of the neurons that do react to inputs is restricted to a narrow range of values leading to poor embedding space statistics.
    The activity statistics of the neurons from the latent space obtained using the binary representation (see Fig.~\ref{fig:BCGAN_latent}) are entirely different. There, no dead neurons can be observed and the level of output values stretches over a much larger range, resulting in a very efficient use of the latent space.
    Although the statistics of the CGAN embedding space appear to be more diverse, this does not translate to better performances as the model is more prone to misclassifying the images with increasing the number of classes.

    \textbf{Sensitivity with respect to the label.}\;
    To visualize the impact of the generated images w.r.t. the label, we set the target label to the Curie temperature $\mathrm{T}_{C}$. In the Ising model, this is  where the phase transition occurs; it is the pivoting point between the occurrence of large, ordered features and purely stochastic features. Any small fluctuation near this temperature should result in a small variation in the generated images while a bigger fluctuation should result in larger changes. For each set of images, the input noise of the GAN is initially fixed, and the same noise is used in all cases. We first generated an image at $\mathrm{T}=\mathrm{T}_C$ and then introduced a small perturbation $\epsilon$ to the conditioning label.
    \begin{figure}[ht!]
        \centering
        \includegraphics[width=\textwidth]{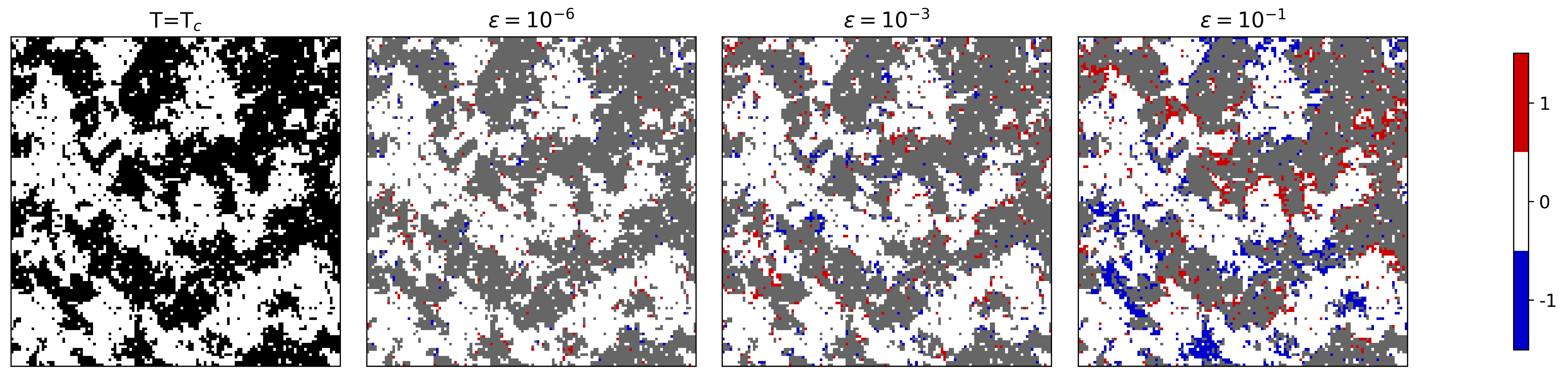}
        \caption{Generated images using the binary embedding space for various values of $\epsilon$ added at $\mathrm{T}=\mathrm{T}_C$. The original microstructure can be seen on the left and the difference between the original image and the generated images with different perturbations are displayed on the three panels to the right. There, the original image (in gray) is superimposed with the modifications (in blue and red).}
        
        \label{fig:ising_small_eps}
    \end{figure}
    From Fig.~\ref{fig:ising_small_eps} we observe for $\epsilon=10^{-6}$ that the generated images are very similar, and only small details change as we are still very close to the targeted temperature.
    As $\epsilon$ increases, the generated microstructure increasingly changes the further away we are from the Curie temperature.
    This behavior is an important requirement for having full control over the property-microstructure relation.
     
\section{Conclusion}   
    By analyzing the embedding space used for conditioning the image generation of various strategies, we showed that the underlying statistic of this space has a strong impact on the model's ability to capture the microstructure-temperature relationship. By introducing a novel embedding strategy, based on the binary representation of floating point numbers, we show that it can produce a rich embedding space capable of conditioning the generated images to the exact desired microstructure property. As creating such microstructures with tailored properties is computationally fast, our model might be one of the building blocks for the accelerated design of materials with microstructures. 

\section*{Acknowledgement} 
    This work was funded by the European Research Council through the ERC Grant Agreement No. 759419 MuDiLingo (“A Multiscale Dislocation Language for Data-Driven Materials”) 


\appendix
\section{Ising Model}\label{appendix:Ising}

The Ising system can be described by its Hamiltonian. The energy of the system is given by the interaction between the neighboring dipoles and the interaction of those dipoles with an external field applied to the system. The Hamiltonian is written as:
\begin{equation}
	\centering
	H = -\sum_{\langle i,j\rangle}J_{ij}\sigma_i\sigma_j - \mu\sum_i h_i \sigma_i
	\label{eq:IsingHamiltonian}
\end{equation}
where $\langle i,j\rangle$ is the sum over the nearest neighbours, $J_{ij}$ is the coupling force between the $i^{th}$ and $j^{th}$ magnetic dipole, $\sigma \in \{-1, 1\}$  is the magnetic dipole of a given site, $\mu$ is the magnetic moment and $h$ is an external field applied to the lattice.

\begin{figure}[htbp]
	\centering
	\includegraphics[width=0.9\textwidth]{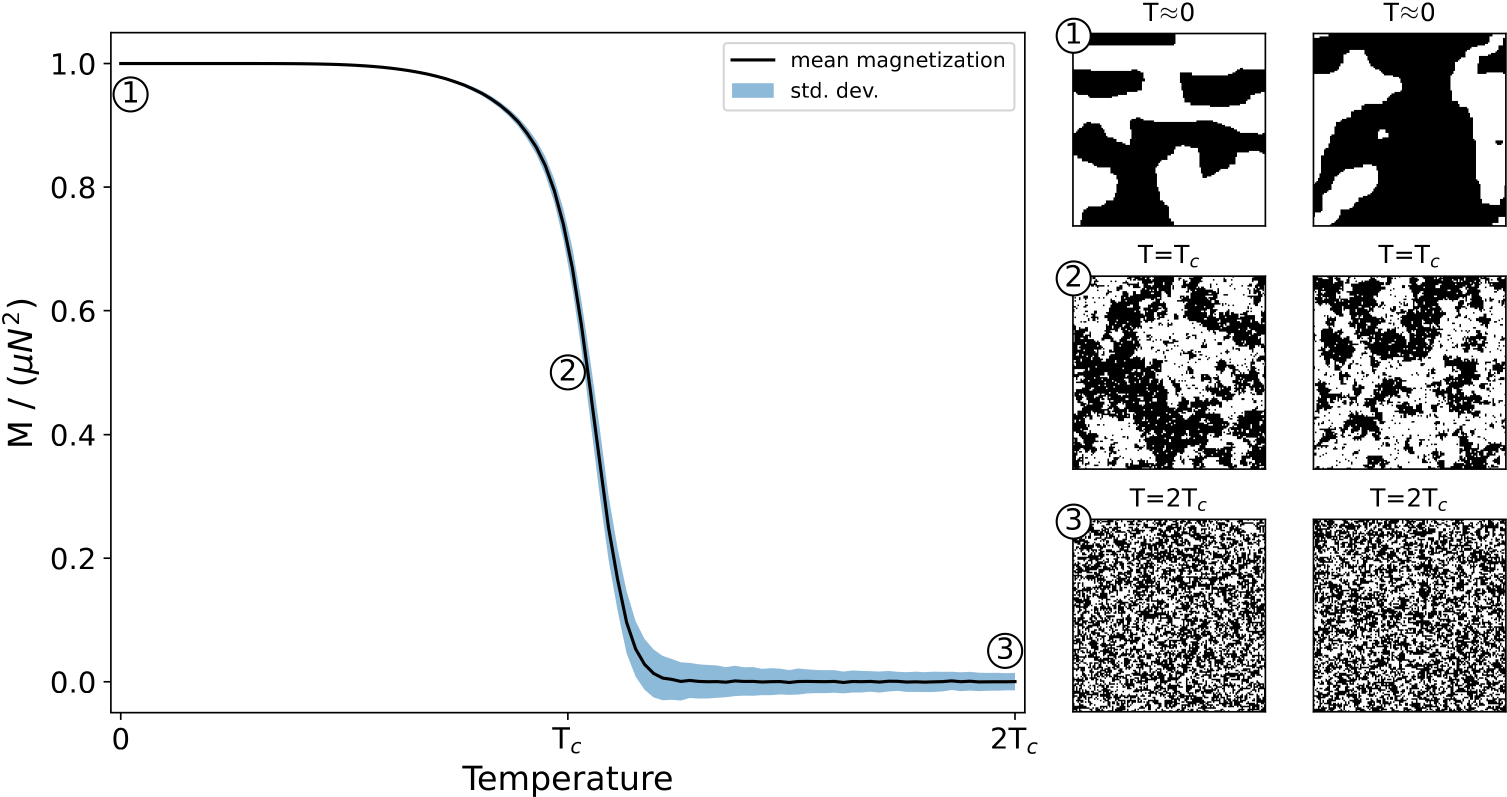}
	\caption{%
		Evolution of the global magnetization of the lattice against the temperature. We can see that the system undertakes a phase transition at the Curie temperature $\mathrm{T}_c$
		Some examples of the microstructure obtained at different temperatures are shown on the right. The 
		image size is $128\times 128$ pixels.
	}
	\label{fig:ising_examples}
\end{figure}
In our case, we simplify Eq.~(\ref{eq:IsingHamiltonian}) by setting the external field $h$ to $0$ and the coupling force $J$ to $1$ for all the magnetic dipole. By fixing $J>0$, we are in the ferromagnetism regime: the spins in the lattice tend to align in the same direction. The equation becomes: 
\begin{equation}
	\centering
	H = -\sum_{\langle i,j\rangle}\sigma_i\sigma_j
	\label{eq:IsingSimple}
\end{equation}

When neighboring dipoles have the same sign, we can see in Eq.~(\ref{eq:IsingSimple}) that the energy of the configuration is minimized.  
One of the specifics of this model is that, in the 2D case and above, the system exhibits a phase transition near the Curie temperature. This phase transition is characterized as going from a large, ordered structure of spins oriented in the same direction, to spins that are randomly oriented without any clear pattern. One way to track this phase transition is with the evolution of the global magnetization $\langle M\rangle$ over a range of temperatures. The global magnetization is the sum of all the dipoles over the number of sites of the lattice. The evolution of the global magnetization and the resulting phase transition can be seen in Fig.~\ref{fig:ising_examples} with the corresponding example of microstructure for different temperatures. Note, that at exactly $T=0$, the whole image should contain only positive or only negative dipoles (which additionally may lead to numerical problems). In the figure, the microstructures for the smallest temperature are therefore obtained for a very small positive value.

\paragraph{Simulation set up}
The Ising system is simulated using the Metropolis Monte Carlo algorithm with periodic boundary conditions. 
We randomly initialize the directions of the  $N \times N$ dipoles in the lattice, where in our case $N=128$ was chosen. Then, we randomly select one site $s(i,j)$, where $s$ stores the lattice state and $i,j = 1,...,N$ and flip the corresponding magnetic dipole. The energy of this new configuration is then calculated with Eq.~(\ref{eq:IsingSimple}). 
If the energy of the new configuration is smaller than the previous one, we keep the new configuration. 
If the energy of this configuration is larger than the previous one, we only keep it with a probability $p=e^{-\beta \Delta E}$, $\Delta E=H_p-H_n$, $H_p$ and $H_n$ is the energy of the previous and next configuration respectively and $\beta = \frac{1}{k_B \mathrm{T}}$, where T is the temperature of the system and $k_B$ the Boltzmann constant, set to $1$ for simplicity.
Those steps are repeated until the system reaches equilibrium or if the number of steps reaches our stopping criterion.
We choose $N^{3}$ as a stopping criterion for the maximum number of steps for MMC algorithm, where N is the size of the lattice. The idea behind this criteria is that every site of the $N \times N$ lattice is visited approximately N times so that the information has enough time to travel through all the lattices.

The dipole configuration from the simulation is then saved as a black and white image
(0 for a negative dipole and 255 for a positive one).
The images are generated from $0$ to $2 \times \mathrm{T}_c$, where $\mathrm{T}_c$ is the Curie temperature at which the system undergoes a phase transition as shown in Fig.~\ref{fig:ising_examples} and is defined as in the general case as

\begin{equation}
	\label{eq:Tc}
	\mathrm{T}_c = \frac{2J}{k_B \ln(1+\sqrt{2} )}
\end{equation}
Which can be simplify to $2/\mathrm{ln}(1+\sqrt{2}) \approx 2.269$ by assuming that $k_B = 1$ and $J=1$.

\section{Architecture Design}\label{appendix:model}

The architectures for CcGAN and BcGAN used the same model, based on the SAGAN architecture \cite{SAGAN}, and were trained for 100 epochs using a batch size of 200. We used the Adam optimizer \cite{Adam} with a learning rate of $10^{-4}$ and $\beta = (0, 0.99)$.

\begin{figure}[htbp]
	\centering
	\includegraphics[width=0.8\textwidth]{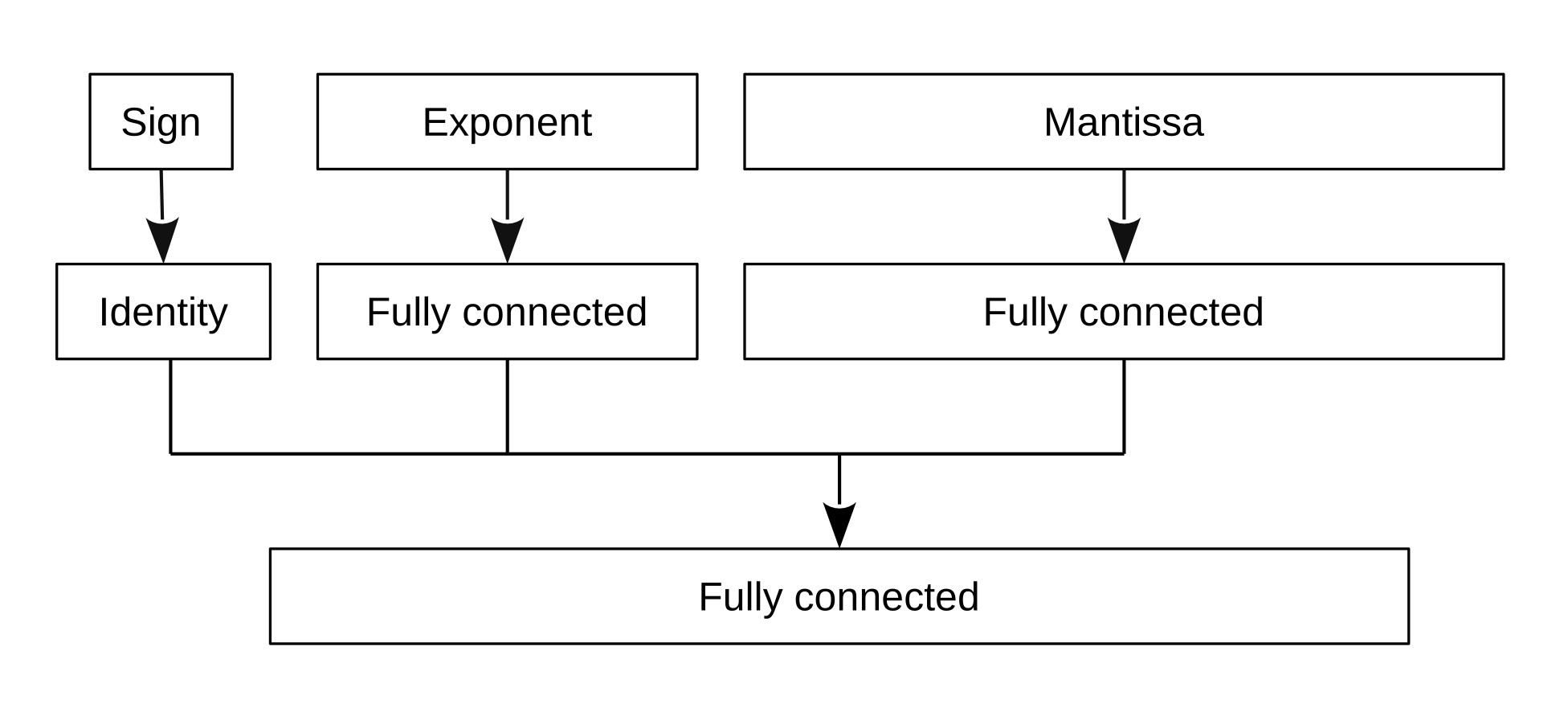}
	\caption{Graphical representation of the embedding network used for the BcGAN architecture}
	\label{fig:embedding}
\end{figure}

While CcGAN normalizes the conditioning values, using the binary representation removes the need for it as all values will be encoded in the same manner, following defined rules. The embedding network consists of two fully connected layers with a hyperbolic activation function used after each layer. The first fully connected layer is divided into 3 parts, one for each constituting part of the binary representation of the floating point number, as seen in Fig.~\ref{fig:embedding}. The embedding network is added inside the generator and the discriminator and is trained alongside the rest of the respective model.

\section{Neuron Statistics} \label{appendix:latent}

A boxplot gives a graphical representation of the distribution of the data, offering a visual summary of its key characteristics. The box in the plot spans the interquartile range, with a line inside representing the median. This provides a quick insight into the central tendency and spread of the data. The whiskers extend from the box, indicating the data variability. 

In order to visualize the statistics from each embedding, each GAN (i.e., the CGAN, CcGAN, and BcGAN) are first trained on the Ising model. Once the models are trained, we extract their embedding layer. Each embedding has a different ``working range'', i.e., the range of numbers they operate with: The CGAN embedding expects a vector containing information on the classes, ranging from 0 to the number of classes in the dataset. The CcGAN embedding expects a vector containing the normalized labels, ranging from 0 to 1. Finally, the BcGAN embedding expects value in the working range of the float32, thus the label of the dataset can directly be used. For each embedding strategy, we give as an input a vector containing values in their respective working range that are uniformly distributed between the minimum and the maximum. As an output, we get the vector used for the conditioning of the model, on which we can compute and display the statistics.

\section{Sensitivity with respect to the label}

\begin{figure}[htbp]\label{appendix:epsilon}
	\centering
	\includegraphics[width=\textwidth]{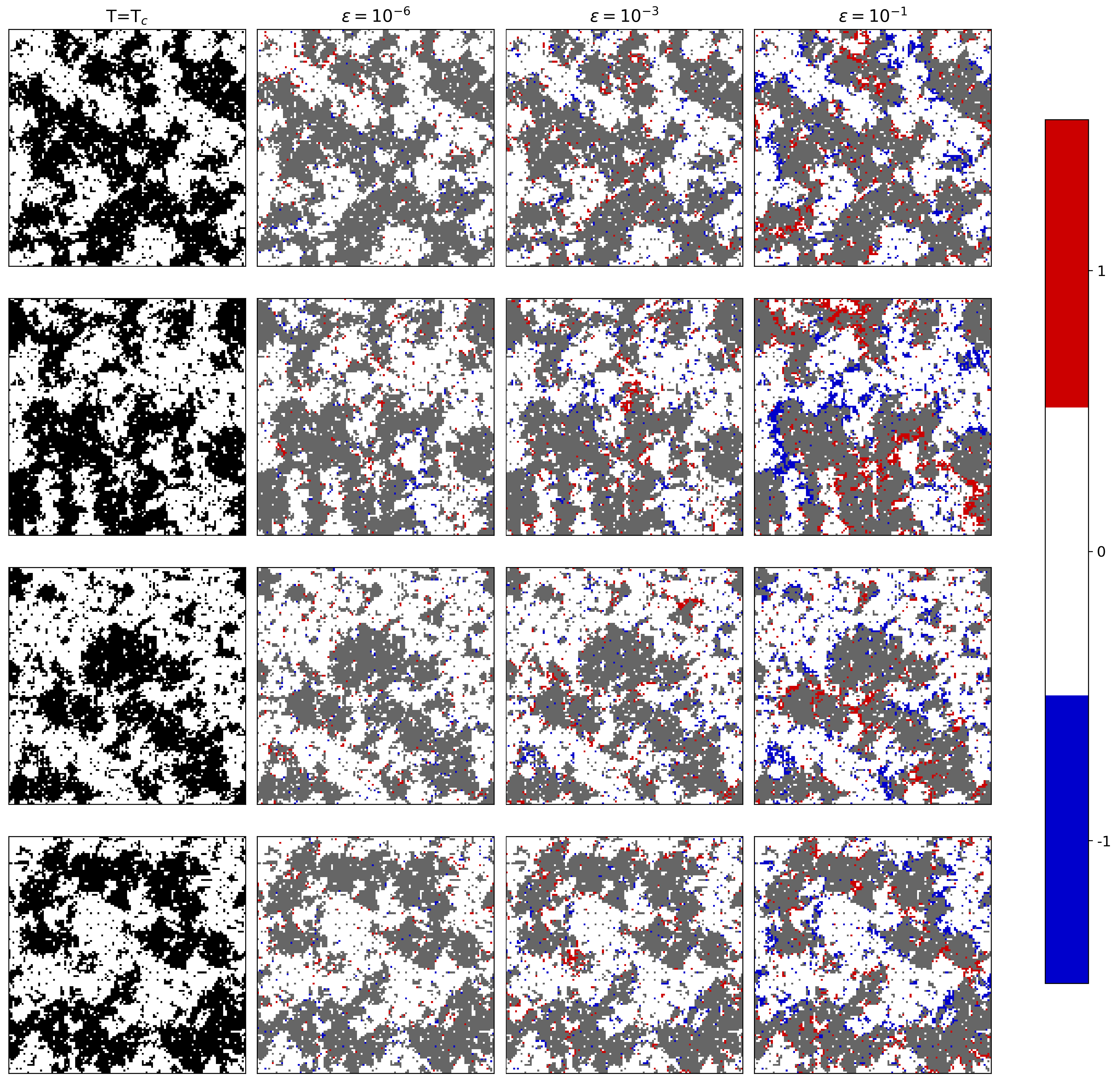}
	\caption{Generated images from BcGAN with different starting noise for various values of $\epsilon$ added at $\mathrm{T}=\mathrm{T}_C$. Each row represents a different noise input for the generator.}
	\label{fig:BCGAN_epsilon_full_64}
\end{figure}

\newpage
\begin{figure}[htbp]
	\centering
	\begin{subfigure}[b]{0.44\textwidth}
		\centering
		\includegraphics[width=\textwidth]{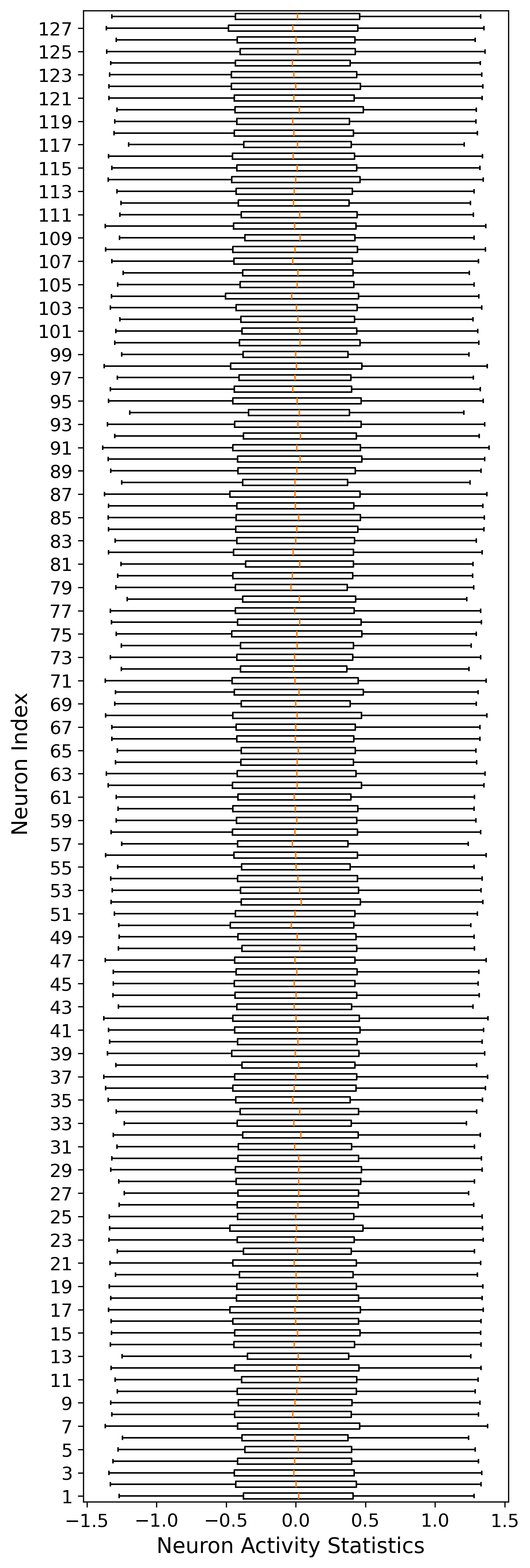}
		\caption{Untruncated boxplot of the activity of the neurons in BcGAN embedding space}
		\label{fig:BCGAN_latent_full}
	\end{subfigure}
	\hfill
	\begin{subfigure}[b]{0.431\textwidth}
		\centering
		\includegraphics[width=\textwidth]{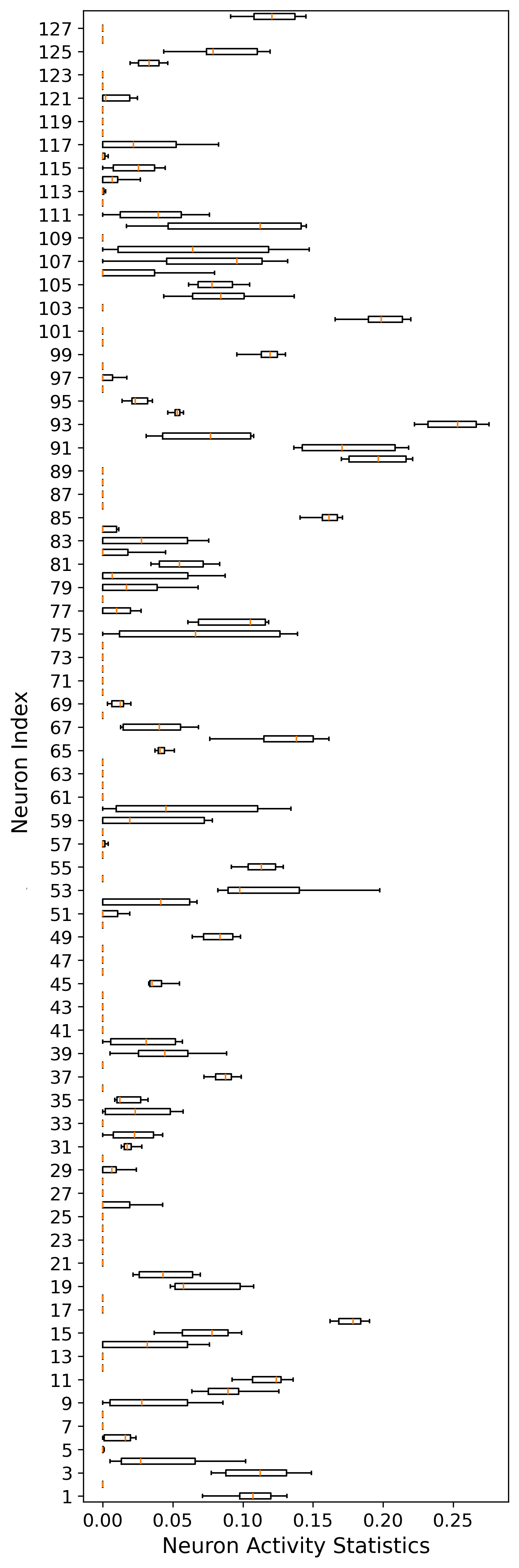}
		\caption{Untruncated boxplot of the activity of the neurons in CcGAN embedding space}
		\label{fig:CCGAN_latent_full}
	\end{subfigure}
	\caption{Comparison of the untruncated neurons activity for both continuous embedding strategies.}
	\label{fig:full_stat}
\end{figure}

\newpage
\begin{figure}[htbp]\label{appendix:lookup}
	\centering
	\begin{subfigure}[b]{0.49\textwidth}
		\centering
		\includegraphics[width=\textwidth]{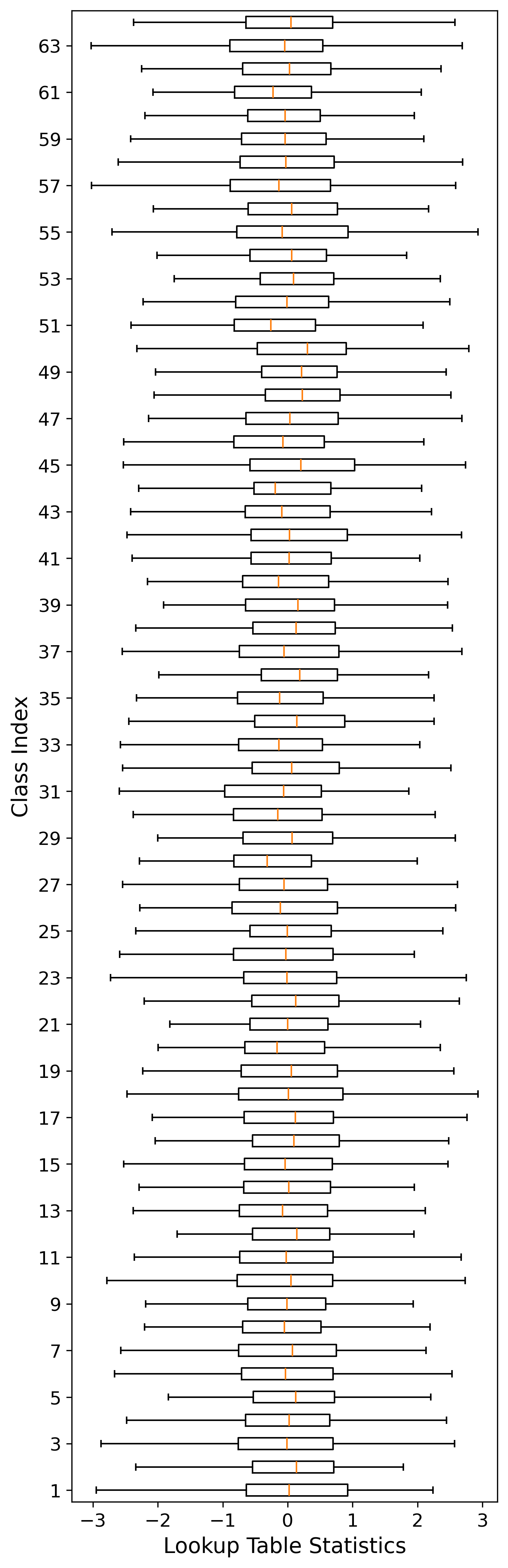}
		\caption{Untruncated boxplot of the lookup table entry for CGAN with 64 classes.}
		\label{fig:CGAN_latent_full_64}
	\end{subfigure}
	\hfill
	\begin{subfigure}[b]{0.46\textwidth}
		\centering
		\includegraphics[width=\textwidth]{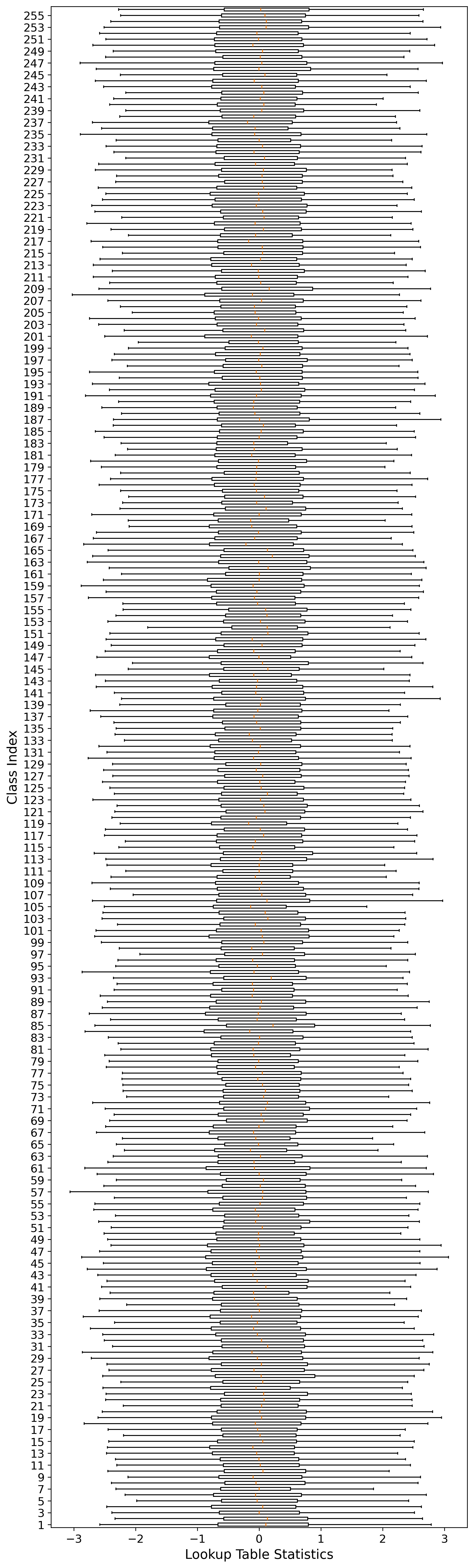}
		\caption{Untruncated boxplot of the lookup table entry for CGAN with 256 classes.}
		\label{fig:CGAN_latent_full_256}
	\end{subfigure}
	\caption{Comparison of the untruncated lookup table of the embedding from the generator of CGAN for two different classes counts. We can see that the statistics stay similar regardless of the number of classes.}
	\label{fig:full_stat_Cgan}
\end{figure}

\end{document}